\documentclass[sigconf]{acmart}

\usepackage{booktabs}
\usepackage{subfig}
\usepackage{bm}
\usepackage{balance}
\usepackage{tikz}
\usepackage{colortbl}
\usepackage{subfig}
\usepackage{graphicx}
\usetikzlibrary{positioning}
\usepackage{enumitem}

\usepackage{amsthm}
\usepackage{amsmath}
\usepackage{amsfonts}
\usepackage{bbm}
\usepackage{chngcntr}
\usepackage{apptools}
\usepackage{algorithm}
\usepackage{algpseudocode}

\DeclareMathOperator*{\argmax}{arg\,max}

\algnewcommand\algorithmicforeach{\textbf{for each}}
\algdef{S}[FOR]{ForEach}[1]{\algorithmicforeach\ #1\ \algorithmicdo}

\newtheorem{definition}{Definition}
\newcommand{\GreedyMask}{\textsc{GreedyMask}}
\newcommand{\RFGreedyMask}{$\textsc{RF}_\textsc{GreedyMask}$}
\newcommand{\MARTGreedyMask}{$\textsc{MART}_\textsc{GreedyMask}$}
\newcommand{\CatCART}{\textsc{CatCart}}
\newcommand{\RFCatCART}{$\textsc{RF}_\textsc{CatCart}$}
\newcommand{\MARTCatCART}{$\textsc{MART}_\textsc{CatCart}$}

\newcommand{\BagOfWords}{\textsc{BagOfWords}}
\newcommand{\PreTrained}{\textsc{PreTrained}}
\newcommand{\Shingling}{\textsc{Shingling}}
\newcommand{\TargetMean}{\textsc{TargetMean}}
\newcommand{\OneHot}{\textsc{OneHot}}

\begin{document}

\title{Modeling Text with Decision Forests using Categorical-Set Splits}

\fancyhead{}

\renewcommand\footnotetextcopyrightpermission[1]{}
\settopmatter{printacmref=false}
\acmConference[]{}{}{}
\fancyhead[LE]{\footnotesize \shorttitle}
\fancyhead[RO]{\footnotesize \shortauthors}
\fancyhead[RE]{\footnotesize \shortauthors}
\fancyhead[LO]{\footnotesize \shorttitle}

\author{Mathieu Guillame-Bert, Sebastian Bruch, Petr Mitrichev, Petr Mikheev, Jan Pfeifer}
\authornote{Correspondence to: Mathieu Guillame-Bert <gbm@google.com>}

\affiliation{%
  \institution{Google}
  \streetaddress{~}
  \city{~} 
  \state{~} 
  \postcode{~}
}
\email{ }

\begin{abstract}
Decision forest algorithms typically model data by learning a binary tree structure recursively where every node splits the feature space into two sub-regions, sending examples into the left or right branch as a result. In axis-aligned decision forests, the ``decision'' to route an input example is the result of the evaluation of a condition on a single dimension in the feature space. Such conditions are learned using efficient, often greedy algorithms that optimize a local loss function. For example, a node's condition may be a threshold function applied to a numerical feature, and its parameter may be learned by sweeping over the set of values available at that node and choosing a threshold that maximizes some measure of purity. Crucially, whether an algorithm exists to learn and evaluate conditions for a feature type determines whether a decision forest algorithm can model that feature type at all. For example, decision forests today cannot consume textual features directly---such features must be transformed to summary statistics instead. In this work, we set out to bridge that gap. We define a condition that is specific to categorical-set features---defined as an unordered set of categorical variables---and present an algorithm to learn it, thereby equipping decision forests with the ability to directly model text, albeit without preserving sequential order. Our algorithm is efficient during training and the resulting conditions are fast to evaluate with our extension of the QuickScorer inference algorithm. Experiments on benchmark text classification datasets demonstrate the utility and effectiveness of our proposal.
\end{abstract}

%
%



\maketitle

\section{Introduction} \label{sec:introduction}

Machine learning algorithms often consume observations in the form of feature vectors in order to learn models. The semantics of each feature defines how it should be consumed by the algorithm and ultimately used in the resulting model. A typical axis-aligned decision tree algorithm, for example, models data by learning a binary tree where each node recursively bifurcates the training examples into sub-regions according to a ``split'' along one of the dimensions of the feature space; what a split looks like and how an optimal split is found for a feature depends entirely on semantics.

A learning algorithms either has the ability to consume a feature type as is or otherwise requires that it be transformed to a supported type. Virtually all algorithms, decision trees included~\cite{breiman1983,jin03,geurts2006}, are able to ingest numerical features, sometimes exclusively so. The record is mixed for other feature types such as graphs, time series, text, or categorical features. Neural networks, for example, require that categorical features be transformed to one-hot vectors or another numerical form. On the other hand, decision forests that use a CART~\cite{breiman1983} split-finding algorithm support categorical features naturally~\cite{catboost,breiman2001,breiman1983}.

Whether or not a feature transformation is used to prepare training data may dramatically affect model training. A contrived transformation step may remove potentially meaningful signals or, conversely, introduce meaningless correlations where none exists. One-hot encoded categorical features, for example, often lead to unbalanced splits in decision trees, leading to sub-optimal models that do not generalize as well as CART-powered decision trees---that is the driving reason for the use of CART in decision forest libraries such as LightGBM~\cite{lightgbm2017nips}. Such empirical observations have motivated researchers to study ways of extending decision trees to support other feature types such as time series~\cite{rodriguez2004,deng2013} and timestamped symbol sequences~\cite{guillamebert2017} among others.

In this work, we set out to enable decision trees to consume another common feature type that remains unsupported to date: categorical sets. A categorical-set feature value is defined as a (by definition, unordered) set of categorical terms. For example, consider a data point that is represented by the following 4 features: $\{ f_1=5, f_2="cat", f_3= \{ "blue", "red", "green" \}, f_4=\emptyset\}$. In this example, $f_1$ is a numerical feature, $f_2$ is a categorical feature, and $f_3$ and $f_4$ are two categorical-set features. Note that, an empty categorical-set feature value is semantically different from a missing value.

A decision tree learner that is equipped with a split-finding algorithm specialized for categorical-set features may naturally consume text, as text can be trivially (albeit incompletely) expressed in that form. Such an extension of decision trees, in turn, allows the application to text corpora of an array of decision forest algorithms such as Random Forest (RF)~\cite{breiman2001}, Multiple Additive Regression Trees (MART)~\cite{friedman2001}, Dropout Multiple Additive Regression Trees (DART)~\cite{rashmi2015}, and Extremely Randomized Trees~\cite{geurts2006}.

Our work formalizes the notion of categorical-set splits and offers a greedy algorithm to learn them efficiently. Our formulation of a categorical-set split tests the presence of any one of a set of terms (called ``mask'') in the feature value set: When the intersection of the mask and feature value is nonempty, the split decision is in the affirmative. This mask itself is learned incrementally using a stochastic, greedy process guided by the decision tree loss function: From a subsample of vocabulary terms, the term that minimizes the loss the most is added to the mask. This process is repeated until the loss cannot be further reduced, at which point the resulting mask is our split.

Our contributions can be summarized as follows:
\begin{itemize}[leftmargin=*]
\item We define and formulate splits (conditions) on categorical-set features in the context of decision trees;
\item We propose an efficient algorithm to learn such splits in a way that is agnostic to the decision tree learner or the local loss function;
\item We report an empirical comparison of our proposed algorithm with methods that require feature transformation;
\item We present an analysis of the stability of the algorithm's hyperparameters; and,
\item We extend the QuickScorer~\cite{lucchese2017} algorithm for efficient inference of models with categorical-set splits.
\end{itemize}

The remainder of this paper is organized as follows. We begin with a brief review of the literature in Section~\ref{sec:related_work}. We present our proposed algorithm in Section~\ref{sec:learning}. Section~\ref{sec:experiments} gives the details of our experimental setup and, in Section~\ref{sec:model_quality} we report a comparison of our proposed method with baselines on a number of publicly available, benchmark datasets. That is followed by a detailed analysis of the resulting models' structure and their sensitivity to the choice of hyperparameters in Section~\ref{sec:analysis}. We turn to efficient model inference in Section~\ref{sec:inference}. We conclude this work in Section~\ref{sec:conclusion} and lay out future directions.
\section{Background and Related Work} \label{sec:related_work}

A decision forest is a collection of decision trees. A decision tree itself is typically a binary tree that routes an example recursively until it reaches a leaf, a terminal node. The decision at every intermediate node to take the left or right branch is, in axis-aligned decision trees, made based on a condition on a single feature. We refer to this condition as a \emph{split}. For example, a split for a numerical feature compares the value of that feature with a threshold---that threshold and the comparison operator together define the split.

When a decision tree is learned, the training algorithm finds the best split (e.g., threshold for a numerical feature) for each node greedily, selecting the split that optimizes a given ``purity'' measure or scoring function such as information gain or Gini index. In the case of numerical features, for example, one may learn a split by sweeping over the values available and choosing a threshold that maximizes information gain. For brevity, we refer to such split finding algorithms as ``splitters.''

The machine learning literature offers many splitters that are suitable for numerical features~\cite{breiman1983,jin03,geurts2006}, categorical features~\cite{catboost,breiman2001,breiman1983}, time series~\cite{rodriguez2004,deng2013}, and timestamped symbol sequences~\cite{guillamebert2017}. For categorical features, for example,~\cite{breiman1983} sorts and iterates over possible values according to the estimated local conditional probability (for classification) or mean (for regression) given the labels. This algorithm is exact in the case of binary classification, but is otherwise approximate. \cite{breiman2001} selects the best split among a set of randomly generated splits. Finally, CatBoost~\cite{catboost} replaces categorical features with (numerical) conditional label statistics. As another example, for time series,~\cite{rodriguez2004} optimizes univariate splits base on Dynamic Time Warping kernels, while both~\cite{rodriguez2004} and~\cite{deng2013} optimize statistics-on-sub-internal type splits. Finally, for timestamped symbol sequences, \cite{guillamebert2017} expresses a split as the matching of a temporal pattern expressed as a graph of time constraints.

To the best of our knowledge, no published work has addressed the challenge of finding splits on categorical-set features---defined in Section~\ref{sec:introduction}. However, several heuristics exist that may be utilized to transform categorical-set features into numerical or categorical values, thereby enabling decision trees to incorporate such features indirectly. We review some of them below.

\textbf{\BagOfWords}, for example, replaces a categorical-set feature with a histogram: the count of occurrences of each term given a vocabulary $D$. More precisely, a feature value $X \subset D$ is replaced with a set of numerical features $\{f_i\}_{i=1}^{|D|}$ with $f_i = | X \cap \{ d_i \} |$ where $d_i \in D$ is a term in the vocabulary. Here we abuse the notion of a set and allow terms to occur repeatedly in $X$.

\textbf{\Shingling}~\cite{manning2008} is a distance between sets of n-grams used in Information Retrieval. Shingling can be used to convert a categorical-set feature $X \subset D$ into a set of numerical features: $\{f_i\}_{i=1}^{k}$ with $f_i = \min_{x \in X} h(x, h_i)$, $h(\cdot)$ a hash function, and $h_i$ a random seed. This approach is related to MinHash~\cite{broder2000Minhash} and Bloom filters~\cite{bloom1970CACM}.

A \textbf{fixed pre-trained representation}, also known as fixed pre-trained embedding, projects every term individually~\cite{mikolov2013,pennington2014}, or the set of terms as a whole~\cite{devlin2019}, into a multi-dimensional dense vector space where terms that are ``close'' in the original representation---for some implicit or explicit definition of closeness---are also close in the target space. Such functions can be learned with a neural network using back-propagation~\cite{devlin2019} or other algorithms~\cite{pennington2014} capable of learning intermediate representations.

A number of recent publications have explored joint training of decision forests and neural networks~\cite{feng2018,balestriero2017,kontschieder2016,bruch2020learning,Ke2019DeepGBM,Li2019CombiningDTandNNs} as a way to harvest the power of deep learning~\cite{lecun2015} to consume text~\cite{devlin2019}, images~\cite{he2016}, graphs~\cite{wang2018} and sets~\cite{zaheer2017}. While not demonstrated, DeepSet~\cite{zaheer2017} is another neural network-based transformation that may be used to incorporate categorical-set features in a decision forest. Note, however, that in this work we are interested in enabling decision trees to consume categorical-set features without transformation of any kind, including representation learning using neural networks.

Another topic that is relevant to the present work is the research on efficient decision forest inference algorithms. Runtime during inference is important because decision forests often comprise of many decision trees, and each decision tree, in turn, requires the evaluation of splits in many intermediate nodes. The cost of traversing paths from a root to leaves and evaluating splits along the way may as such become a bottleneck.

The inference cost is especially high if one na\"ively evaluates a decision tree on an input example: Start at the root and evaluate its condition in order to route the example to one of its (left or right) branches, then repeat that operation until a leaf is reached~\cite{asadi2014}. Though trivial to implement, that approach is suboptimal. Researchers have thus developed more optimized inference algorithms such as QuickScorer~\cite{lucchese2017} (QS) and its extensions V-QuickScorer~\cite{lucchese2016} (v-QS) and RapidScorer~\cite{ye2018}. The core idea there is to evaluate a decision tree by evaluating all its nodes simultaneously, and subsequently retrieving the ``active'' leaf. Despite having an exponentially higher worse-case time complexity, these methods run orders of magnitude faster than the top-down approach on modern CPUs because of their more predictable memory access pattern and branching. In this work, we extend QuickScorer to categorical-set splits.
\section{Categorical-Set Splits} \label{sec:learning}

In this section, we formally define categorical-set splits, present an algorithm to learn such splits, and describe how such splits may be efficiently evaluated during inference. But first, we begin by laying out a set of constraints.

\subsection{Constraints and Considerations}
\label{sec:considerations}
Decision forest learning algorithms such as RF~\cite{breiman2001}, MART~\cite{friedman2001}, or DART~\cite{rashmi2015} all rely on a splitter subroutine to find an optimal split for every node in the tree. During training, splitters are invoked to search a large space of feature values given a potentially large number of training examples in order to arrive at a condition that optimizes a scoring function. This procedure is repeated for every feature and every node independently. During evaluation or inference, the resulting split is computed at every intermediate node in every decision tree to determine which branch an example should be routed towards, often under tight latency constraints~\cite{lucchese2017,lucchese2016,asadi2014}. As such, splits and splitters play an outsize role in the efficiency of the training and inference procedures. It is therefore imperative that any proposed split and splitter be computationally cheap.

Splitters also affect the generalizability of decision forest models: an overzealous splitter that produces a split that overfits the training data leads to poor generalization. This behavior can be controlled with regularization such as by using a regularized loss term, bagging, and training decision trees on a small subsample of examples or features~\cite{breiman1996,xgboost2016KDD,lightgbm2017nips,ganjisaffar:2011:bagging}. The same ideas can be utilized to afford a certain degree of stochasticity to splitters to prevent overfitting. For example, presenting a splitter with only a subsample of training examples or of feature values is one way to introduce uncertainty. We incorporate the latter approach in our proposed algorithm.

\begin{algorithm}[t]
  \caption{\textsc{GreedyMask} algorithm for finding categorical-set splits}\label{algo:splitter}
  \flushleft
  \hspace*{\algorithmicindent}
  \textbf{Input:} Collection of categorical-set features $\mathcal{X}$ and labels $\mathcal{Y}$. Sampling rate $p \in (0,1]$. Vocabulary $D = \{d_j\}_{j=1}^{m}$.\\
  \hspace*{\algorithmicindent}
  \textbf{Output:} Split mask $M$. 
  \begin{algorithmic}[1]
    \State $\hat{D} \gets \{ d \mid d \in D \text{ with probability} \; p \}$
    \State $M \gets \emptyset$ \Comment{Initial empty mask}
    \While{$\textbf{true}$}
        \State $\hat{d} \gets \argmax_{d \in \hat{D}} \textsc{score} (\mathcal{X}, \mathcal{Y}, M \cup {d})$
        \If{$\textsc{score}(\mathcal{X}, \mathcal{Y}, M \cup {\hat{d}}) \leq 0$}
            \State \textbf{break}  \Comment{No further improvement possible}
        \EndIf
        \State $\hat{D} \gets \hat{D} \setminus {\hat{d}}$
    \EndWhile
  \end{algorithmic}
\end{algorithm}

\subsection{Definition and Learning}

Algorithm~\ref{algo:splitter} presents our proposed splitter for categorical-set features. To understand the algorithm, let us define its output first: a categorical-set split. Like splits on numerical features that compare a value with a threshold, a categorical-set split consists of an operator and a ``threshold.'' We choose intersection as the operator and a fixed subset of the vocabulary as the threshold, or more appropriately, the mask. A split on a categorical-set feature is a test of whether its intersection with the mask is nonempty. The following definition formalizes this concept.
\begin{definition}
Given a vocabulary of possible terms $D$ and a categorical-set feature $X \subset D$, a Categorical-Set Split is the result of $X \cap M \neq \emptyset$, where $M \subset D$, a \emph{mask}, is a fixed set of terms associated with the split.
\end{definition}

The objective of the splitter is then to learn a mask $M \subset D$ such that the split formulated above optimizes a scoring function on a given set of training examples---a list of $n$ categorical-set feature values $\mathcal{X} = \{X_i\}_{i=1}^{n}$ with $X_i \subset D$, and their corresponding labels $\mathcal{Y} = \{y_i\}_{i=1}^{n}$. We note that any scoring function may be used but typical examples are information gain for classification with Random Forests, and mean squared error for Gradient Boosted Decision Trees.

Let us now describe how Algorithm~\ref{algo:splitter} constructs $M$. It first samples a random subset $\hat{D} \subset D$ with each term sampled independently with probability $p$, a hyperparameter that controls the stochasticity of the split. This step follows from the discussion in Section~\ref{sec:considerations} on the impact of splitters on generalization. In effect, through sampling, we aim to prevent the splitter from incorporating the entire vocabulary in a mask, as otherwise the resulting mask would severely overfit the training data. We examine the effect of hyperparameter $p$ later in this study.

Now that we have down-sampled $D$ into $\hat{D}$, we initialize $M$ to the empty set and proceed iteratively as follows: In each iteration, the algorithm finds the vocabulary term from $\hat{D}$ that maximizes an arbitrary scoring function ($\textsc{score}(\cdot)$). That term is then added to the mask, $M$. This procedure continues until no additional vocabulary term improves the score.

\subsection{Efficient Inference}
In this section, we show how an extension of QuickScorer~\cite{lucchese2017} (QS) may support splits learned by Algorithm~\ref{algo:splitter}, thereby facilitating efficient inference. To understand our proposed extension, let us briefly review the mechanism by which QS determines which leaf is active given an input example with numerical features: QS begins by constructing a ``leaf mask,'' a bit vector, initially all set, whose size is equal to the number of leaves in the tree. Each node too has a ``node mask,'' a bit vector of the same size that encodes the leaves that are unreachable if an example fails to satisfy that node's split condition. QS proceeds by taking one numerical feature at a time and iterating over all nodes that split on that feature. If the node's threshold is smaller than the feature value, that node's mask is applied (with a bitwise \texttt{AND}) to unset unattainable leaves. In the end, the index of the lowest set bit in the leaf mask is the active leaf. Note that, a separate leaf mask is maintained per tree.

\begin{figure}
\begin{center}
\includegraphics[width=0.9\linewidth]{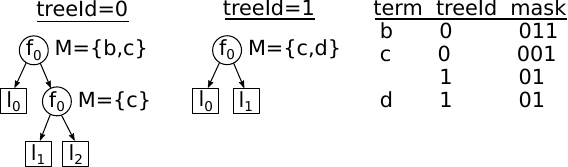}
\caption{
Illustration of \texttt{termMask} for an example forest with a single categorical-set feature $f_0$ with vocabulary $\{a, b, c, d\}$. If an example feature contains the term \emph{c}, leaves $l_0$ and $l_1$ are not reachable in the first tree, resulting in the term mask $001$. Similarly, leaf $l_0$ from the second tree is not reachable, leading to the mask $01$. The presence of \emph{a} is inconsequential for the term mask of both trees. So is the presence of \emph{b} for the second tree's term mask and \emph{d} for the first tree's.
}
\label{fig:quickscorerexemple}
\end{center}
\end{figure}

Algorithm~\ref{algo:inference} presents our extension of QS. We adopt the same notation as in Algorithm 1 in the original work~\cite{lucchese2017} and refer the reader to that work for a complete account. For our algorithm to work, we prepare the following data structure for each categorical-set feature separately: We compile what we refer to as ``term masks,'' bit vectors that are similar to node masks in QS but that encode, for each term in the vocabulary, the leaves that are unreachable if an example contains that term. Figure~\ref{fig:quickscorerexemple} shows an example decision forest with categorical-set splits along with its term masks.

Once term masks are built for all categorical-set features and all trees in the forest, we group them by feature and sort each group by term into an array. This is the \texttt{termMask} structure in Algorithm~\ref{algo:inference}. Note that, in each group, a term may have more than one term mask as it may appear in splits in more than one decision tree. Finally, for each group, we index every term by storing its start and end indices in \texttt{termMask}. This organization results in a compact and access-efficient representation.

\begin{algorithm}[t]
  \caption{Extension of QuickScorer for categorical-set splits. Notation and surrounding code follow Algorithm 1 in~\cite{lucchese2017}.}\label{algo:inference}
  \flushleft
  \hspace*{\algorithmicindent}
  \textbf{Input:} An input example $x$.
  \hspace*{\algorithmicindent}
  \begin{algorithmic}[1]
    \State{Init leaf mask, \texttt{leafidx}, indexed by trees} \Comment{Lines 1--3 of~\citep{lucchese2017}}
    \State{Apply numerical node masks to leaf masks} \Comment{Lines 4--9 of~\citep{lucchese2017}}
    
    \ForEach {Categorical-Set Feature $f$}
        \ForEach {$v \in x[f] $} \Comment{Terms present in feature $f$ of $x$}
            \State $r \gets \texttt{termMaskIndex} [f][v]$
            \ForEach {$w \in [r.begin, r.end ) $}
                \State $m \gets \texttt{termMask}[f][w]$
                \State $\texttt{leafidx} [ m.treeId ]\,\wedge{=} m.mask$
            \EndFor
        \EndFor
    \EndFor
    
    \State{Retrieve leaf indices} \Comment{Lines 10--15 of~\cite{lucchese2017}}
  \end{algorithmic}
\end{algorithm}

\section{Experimental Setup} \label{sec:experiments}

This section reports the empirical evaluation of our proposed splitter algorithm on RF and MART algorithms on 5 public text classification datasets. We begin with a description of these datasets, list the methods under evaluation, and finally present and discuss the results.

\subsection{Datasets}

We consider 5 binary classification datasets of the cleaned Sentiment Analysis dataset repository~\cite{senteval}. Table~\ref{tab:datasets} shows the names and statistics of our datasets. We tokenize the text features by white space, thereby representing each piece of text as a set of unigrams. Once classifiers are trained, we measure the Area Under the Receiver Operating Characteristic Curves (AUC), averaged in a 5-fold cross-validation scheme.

\begin{table}
\caption{Dataset statistics. Average number of terms by example shows the average size of deduplicated categorical-sets.}
\label{tab:datasets}
\begin{tabular}{@{}lrrr@{}}
\toprule
Dataset & \#Examples & \#terms/examples \\ \midrule
Stf. sentiment treebank (SST) & 68.8k & 9.8 \\
Product review (CR) & 8k & 20.1 \\
Movie review (MR) & 22k & 21.6 \\
Subjectivity status (SUBJ) & 20k & 24.6 \\
Opinion-polarity (MPQA) & 22k & 3.1 \\ \bottomrule
\end{tabular}
\end{table}

\subsection{Methods}
We consider several learning algorithms including Neural Networks (NN), Linear classifier (Linear), Random Forests (RF) and Multiple Additive Regression Trees (MART)---the last two being decision forest algorithms. In order to evaluate our proposal we make two measurements. In one, we measure the effectiveness of our categorical-set splits for decision forests, applied directly to the datasets. The second trains a model using the learning algorithms above with text features transformed using one of the following functions:
\begin{itemize}[leftmargin=*]
\item \textbf{\TargetMean}, inspired by CatBoost~\cite{catboost}, replaces a categorical feature by the conditional label distribution of its values, estimated on the training set. For example, in a binary classification setting, the categorical value ``A'' is replaced by the ratio of positive labels among examples with value ``A'';

\item \textbf{\Shingling} as described in section~\ref{sec:related_work};

\item \textbf{\BagOfWords} as described in section~\ref{sec:related_work};

\item \textbf{\OneHot}, reduces \BagOfWords{} to categorical features; and,

\item \textbf{\PreTrained} is a 128-dimension term based text embedding~\cite{bengio2003} trained on the English Google News 200B corpus.~\footnote{Available at https://tfhub.dev/google/nnlm-en-dim128/1}
\end{itemize}

In the sections that follow, we adopt the following naming format: Method names begin with the learning algorithm (e.g., RF) followed by a sequence of pre-processing steps (if any) separated by the plus sign. For example, ``RF \Shingling + \TargetMean'' indicates that a Random Forest model is trained where the raw text features were transformed using \Shingling{} first, followed by \TargetMean{}. \CatCART{} indicates that a categorical feature is consumed with the CART splitter~\cite{breiman1983}. Finally, our proposed method is denoted by \GreedyMask{}.

Many of the hyperparameters we used in this work are set to reasonable default values guided by previous publications~\cite{lightgbm2017nips,xgboost2016KDD,catboost}, while a subset (e.g., vocabulary size, sampling rate) are determined by a small-scale validation and fixed across experiments. The following provides a summary:
\begin{itemize}[leftmargin=*]
\item Tokenization: We keep the 5000 most frequent terms that appear at least $5$ times. This is computed independently on the training partition of each cross-validation iteration.
\item RF: We train 500 trees with a maximum depth of 32; the number of features randomly chosen to find a split in a node is the square root of the total number of features.
\item MART: shrinkage is set to $.1$; maximum depth is 6 and number of trees is set to 500 with early stopping using $10\%$ of the training dataset as validation; feature subsampling is disabled; and we use exact splitting for numerical features.
\item NN: 3 layers with 32 units each; batch size is 32; train for a maximum of 20 epochs; early stopping using 10\% of the training dataset as validation; finally, we use the AdaGrad~\cite{Duchi2011AdaGrad} optimizer.
\item Linear: 32 examples per batch; train for 20 epochs with the AdaGrad optimizer.
\item \GreedyMask{}: sampling rate of $.2$. We provide an analysis of the effect of this hyperparameter in Section~\ref{sec:hp_stability}.
\end{itemize}

\section{Model Quality} \label{sec:model_quality}

Table~\ref{tab:aucs} shows the AUCs of the methods under consideration (in rows) on all datasets (in columns), averaged over 5-fold cross-validation trials. We also report the mean and median rank of each method in the same table.

\begin{table*}
\caption{Five-fold cross-validation AUC of the various methods on the 5 binary classification datasets. The mean AUC is accompanied by the standard deviation and, in parentheses, the rank of the method for each dataset. Our methods are bold-faced.}
\tabcolsep=.11cm
\small
\begin{tabular}{@{}lrrrrrrr@{}}
\toprule
\textbf{Method} & \textbf{Median Rank} & \textbf{Avg Rank} & \textbf{SST} & \textbf{MR} & \textbf{CR} & \textbf{MPQA} & \textbf{SUBJ} \\  \midrule
\textbf{\RFGreedyMask} & 1 & 3.8 & \textbf{.9636$\pm$.0039 (1)} & \textbf{.842$\pm$.0458 (1)} & \textbf{.8723$\pm$.0448 (1)} & .8432$\pm$.0181 (13) & .9673$\pm$.0104 (3) \\
MART \BagOfWords & 3 & 4.6 & .9561$\pm$.00556 (3) & .8351$\pm$.0491 (2) & .8559$\pm$.0439 (3) & .8384$\pm$.0157 (14) & \textbf{.9691$\pm$.0094 (1)} \\
\textbf{\MARTGreedyMask} & 3 & 5.2 & .958$\pm$.00399 (2) & .8327$\pm$.0522 (3) & .8522$\pm$.0483 (4) & .8374$\pm$.0134 (15) & .9681$\pm$.0092 (2) \\
Linear \PreTrained & 6 & 6.4 & .9187$\pm$.0122 (15) & .8213$\pm$.0157 (8) & .8652$\pm$.0186 (2) & \textbf{.9342$\pm$.0115 (1)} & .9643$\pm$.013 (6) \\
RF \Shingling+\TargetMean & 7 & 6.8 & .9407$\pm$.00766 (9) & .8302$\pm$.014 (4) & .8439$\pm$.0447 (7) & .8887$\pm$.0104 (6) & .9633$\pm$.013 (8) \\
MART \Shingling+\TargetMean & 7 & 8 & .9375$\pm$.00764 (11) & .8293$\pm$.0188 (6) & .8305$\pm$.0293 (11) & .8897$\pm$.00715 (5) & .964$\pm$.0114 (7) \\
RF \PreTrained & 9 & 9 & .9482$\pm$.00763 (7) & .8023$\pm$.0298 (13) & .8423$\pm$.0324 (9) & .9278$\pm$.00954 (3) & .9458$\pm$.0202 (13) \\
NN \PreTrained & 9 & 9.4 & .9129$\pm$.0113 (16) & .7992$\pm$.0286 (14) & .8466$\pm$.0235 (6) & .9324$\pm$.0188 (2) & .963$\pm$.0149 (9) \\
Linear \Shingling+\TargetMean & 9 & 9.8 & .8991$\pm$.0102 (17) & .8294$\pm$.0179 (5) & .7713$\pm$.121 (14) & .8634$\pm$.0121 (9) & .9662$\pm$.0106 (4) \\
MART \PreTrained & 10 & 9.4 & .938$\pm$.00842 (10) & .8056$\pm$.0313 (11) & .8299$\pm$.0532 (12) & .9259$\pm$.0164 (4) & .9568$\pm$.0148 (10) \\
Linear \Shingling+\OneHot & 10 & 11 & .9448$\pm$.00652 (8) & .8062$\pm$.0205 (10) & .7113$\pm$.471 (15) & .8763$\pm$.0114 (7) & .7953$\pm$.626 (15) \\
NN \BagOfWords & 11 & 10 & .9308$\pm$.00643 (14) & .8073$\pm$.0384 (9) & .8486$\pm$.0641 (5) & .8578$\pm$.0195 (11) & .9539$\pm$.0131 (11) \\
Linear \BagOfWords & 12 & 12 & .937$\pm$.00577 (12) & .7907$\pm$.0338 (18) & .8437$\pm$.054 (8) & .8593$\pm$.0169 (10) & .9469$\pm$.0193 (12) \\
NN \Shingling+\TargetMean & 12 & 11 & .8921$\pm$.0112 (18) & .8279$\pm$.0175 (7) & .8146$\pm$.0604 (13) & .8448$\pm$.0194 (12) & .9651$\pm$.0119 (5) \\
NN \Shingling+\OneHot & 13 & 13 & .9365$\pm$.00815 (13) & .8042$\pm$.0284 (12) & .7064$\pm$.463 (16) & .8724$\pm$.0153 (8) & .7905$\pm$.63 (16) \\
\RFCatCART \Shingling & 17 & 14.6 & .9535$\pm$.00495 (4) & .7913$\pm$.0167 (17) & .696$\pm$.459 (17) & .8246$\pm$.0152 (16) & .7717$\pm$.687 (19) \\
\MARTCatCART \Shingling & 17.5 & 15.1 & .9512$\pm$.00447 (5.5) & .7922$\pm$.0271 (15.5) & .6707$\pm$.433 (19.5) & .822$\pm$.0127 (17.5) & .7735$\pm$.694 (17.5) \\
MART \Shingling+\OneHot & 17.5 & 15.1 & .9512$\pm$.00447 (5.5) & .7922$\pm$.0271 (15.5) & .6707$\pm$.433 (19.5) & .822$\pm$.0127 (17.5) & .7735$\pm$.694 (17.5) \\
RF \BagOfWords & 19 & 16.4 & .8063$\pm$.0072 (19) & .7416$\pm$.0405 (19) & .8309$\pm$.0268 (10) & .7365$\pm$.042 (20) & .9119$\pm$.0272 (14) \\
RF \Shingling+\OneHot & 20 & 19.4 & .7794$\pm$.0219 (20) & .7166$\pm$.0226 (20) & .6864$\pm$.394 (18) & .7453$\pm$.0408 (19) & .7525$\pm$.517 (20)  \\ \bottomrule
\end{tabular}
\label{tab:aucs}
\end{table*}

The results in Table~\ref{tab:aucs} show that our proposed algorithm when applied to Random Forests leads to considerable gains: The method comes first in terms of median rank, and is the best performing method on 3 of the 5 datasets. However, the method ranks poorly (13/20) on the MPQA dataset, falling far behind Linear \PreTrained{}. We believe this unusual gap is an artifact of the dataset itself: Sentences in MPQA are very short, rarely exceeding a handful of terms---the average number of terms per example, as shown in Table~\ref{tab:datasets}, is a measly $3.1$. With so few terms, it is easy for \CatCART{} and \GreedyMask{} to overfit. \PreTrained{}, in contrast, is at an advantage as its representations are learned using another, larger dataset, making it less prone to overfitting.

As anticipated, the impact of pre-trained embeddings depends on the dataset. Pre-trained embeddings perform well on the MPQA dataset---the top four approaches use embeddings---whereas on other datasets the advantage is somewhat limited. Interestingly, linear models appear to yield higher AUCs when trained on pre-trained embeddings.

$\GreedyMask$ performs better with RF than with MART. This can be partially explained by the stochasticity in Random Forests: On datasets with a low example-to-feature ratio, RFs have low variance as individual decision trees only model a subsample of features. \GreedyMask{} creates a large feature space, leading to effects similar to presenting the algorithms with a large number of features.

\begin{figure}[b]
\begin{center}
\includegraphics[width=\linewidth]{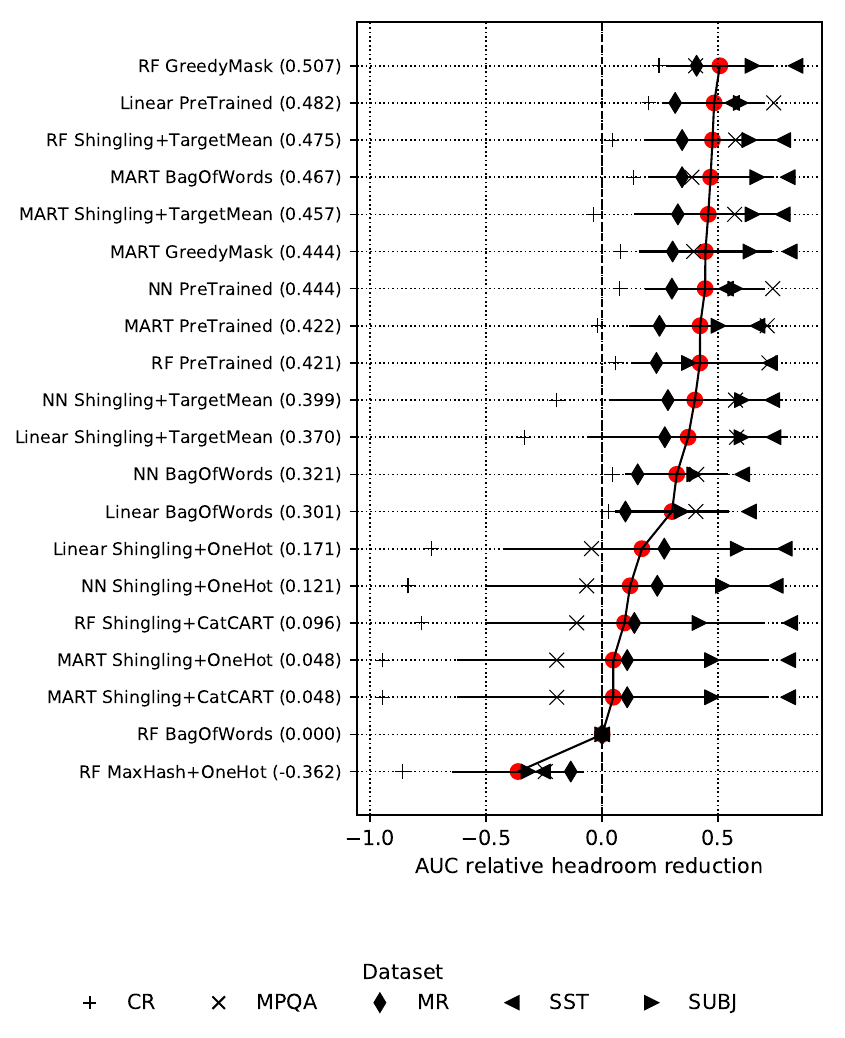}
\caption{Relative accuracy headroom reduction (RAHR) of methods on 5 datasets with RF \BagOfWords\ serving as the baseline. The red circles traced across methods are mean RAHR of each method and the solid horizontal line around each circle represents one standard deviation in both directions.}
\label{fig:rahrs}
\end{center}
\end{figure}

We note that, the ranking above does not take into account the differences in AUC between methods: Small differences matter as much as large ones. A more appropriate comparison would be to measure the gain towards the optimal AUC of 1 relative to a fixed baseline: $\frac{\mathrm{accuracy} - \mathrm{baseline}}{ 1 - \mathrm{baseline}}$. We call this quantity the \emph{relative accuracy headroom reduction} (RAHR) between methods and use RF \BagOfWords{} as baseline. This statistic also aids in the visualization of the results of our experiments by highlighting relative gains, as Figure~\ref{fig:rahrs} illustrates.

Our proposed \GreedyMask{} with RF has an RAHR of $.507$, followed by Linear \PreTrained{} (RAHR=$0.482$) and RF \Shingling+\TargetMean{} (RAHR=$.475$). The overall order of RAHR and ranks are similar, with the main difference being between the top contenders: Linear \PreTrained{} and RF \Shingling+\TargetMean{} go from global ranks 4 and 5, respectively, to ranks 2 and 3 largely due to the greater relative gain of these methods on the MPQA dataset.

\section{Model Analysis} \label{sec:analysis}
In this section, we take a closer look at the methods considered in this work. We begin with a comparison of the structure of the learned models. We then examine the effect of hyperparameters on model performance.

\subsection{Structure}
Table~\ref{fig:model_stats} reports model statistics resulting from the utilization of different pre-processing transformations with the RF algorithm on two datasets. We note that similar conclusions can be drawn from the other 3 datasets, which we have omitted for brevity.

We observe that \BagOfWords{} and \Shingling+\OneHot{} lead to deeper trees, while other solutions learn much shallower trees. One possible interpretation is that node splits resulting from features transformed using \BagOfWords{} and \Shingling+\OneHot{} afford little separability powers, and as a consequence, more splits are required to obtain better decision boundaries. It is also worth noting that \BagOfWords{} effectively tests one term at a time, leading to larger trees that generalize poorly.

It does not come as a surprise then that \BagOfWords{} and \Shingling+\OneHot{} have significantly smaller balance ratios relative to other methods, indicating that trees are on average less balanced. This phenomenon too can be explained by the fact that splits consider a single term (or a single random hash) at a time, thereby repeatedly forcing training examples down the negative branch, ultimately resulting in unbalanced trees.

\begin{table}
\caption{Structure statistics of the RF models for the MR and SST datasets. Balance ratio is the ratio of the log of the number of nodes per tree ($\log_2(\#Nodes/Tree)$) to the average depth---a balance ratio of $1$ indicates a fully balanced tree. } 
\tabcolsep=.11cm
\label{fig:model_stats}
\begin{tabular}{@{}lrr|rr|rr@{}}
\toprule
& \multicolumn{2}{c|}{Avg Depth} & \multicolumn{2}{c|}{\#Nodes/Tree} & \multicolumn{2}{c}{Balance} \\
Method & MR & SST & MR & SST & MR & SST \\
\midrule
\textbf{\RFGreedyMask} & 11.5 & 17.5 & 477 & 3499 & .771 & .671 \\
RF \BagOfWords & 20.5 & 21.6 & 815 & 1642 & .472 & .494 \\
\RFCatCART Shingling & 12.0 & 12.6 & 369 & 855 & .713 & .775 \\
RF {\small \Shingling+\TargetMean} & 13.5 & 17.0 & 1493 & 6109 & .782 & .739 \\
RF \Shingling+\OneHot & 19.9 & 20.1 & 722 & 1099 & .477 & .503 \\
\bottomrule
\end{tabular}
\end{table}

\subsection{Hyperparameter Stability} \label{sec:hp_stability}
Our proposed method has a single hyperparameter, a sampling rate $p$, which introduces randomness in the splitter. By incorporating this hyperparameter, we hoped to allow a form of regularization and prevent overfitting. In this section, we study the effect of the sampling rate on the final model across different datasets.

\begin{figure}
\begin{center}
\includegraphics[width=1.0\linewidth]{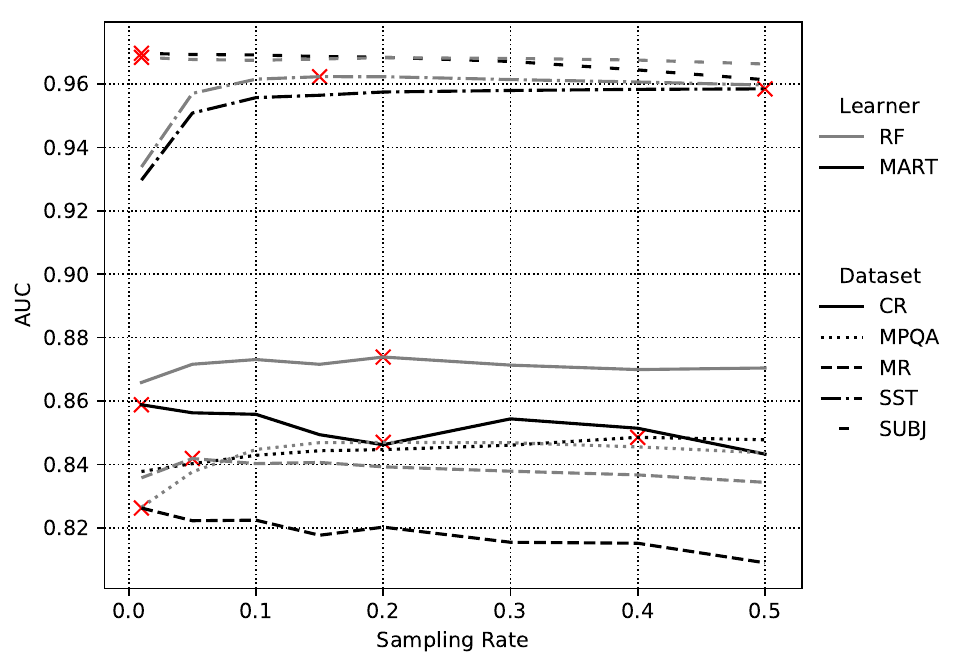}
\caption{
Mean AUC with respect to sampling rate. The cross represents the highest AUC for each method-dataset pair.
}
\label{fig:stab}
\end{center}
\end{figure}

Figure~\ref{fig:stab} shows the change in mean AUC for different values of sampling rate. As before, AUCs are estimated with 5-fold cross-validation. Other hyperparameters are left unchanged (see Section~\ref{sec:experiments}), with the exception of the vocabulary size which is adjusted from 5000 to 2000 terms to facilitate faster experiments.

Model performance is relatively stable and does not change dramatically with changes in the sampling rate: Excluding the sampling rate of $.01$, the average difference between the best and worst AUCs for $p \in [0.05,0.5]$ is only $.0078$. The optimal sampling rate naturally depends on the dataset, ranging from the smallest to the largest tested values. While not reported in Figure~\ref{fig:stab}, for some datasets the optimal sampling rate appears to be $1.0$, meaning no sampling at all. On average, however, $p=.2$ is a reasonable default value for RF with an average $.0026$ AUC drop from the best setting.

Confirming the results of Section~\ref{sec:experiments}, RF performs better than MART by an average 0.0063 in AUC. By construction, RF is less prone to overfitting than MART and, as such, can better correct our splitting algorithm's tendency to overfit to the training data. More work, however, is required to understand and improve our proposed solution for use with MART.

\section{Efficient Model Inference} \label{sec:inference}

We put the extended QS in Algorithm~\ref{algo:inference} to the test. Table~\ref{table:inference_speed} reports its inference speed on the first fold of experiments in Section~\ref{sec:experiments}, and compares it with the direct top-down approach, VPred\cite{asadi2014}. This benchmark considers only the application of a trained model on already-processed input features; in other words, we exclude the time required to perform tokenization or compute hashes or label statistics. The experiments are run with a single thread on a 3.70GHz Intel Xeon CPU. For VPred, both the model mask and the examples are sorted prior to the benchmark in order to evaluate the intersection condition in linear time with the number of items. The results are averaged over 100 runs over the entire dataset, and are preceded by 10 warm-up runs. Numerical splits are evaluated with SIMD instructions (v-QS). The categorical-set split evaluation does not rely on SIMD instructions.

The QS implementation for categorical-set split is nearly 13x faster than the VPred implementation, demonstrating that such splits are well-suited for the QS algorithm. \GreedyMask{} (withour SIMD instructions) runs nearly 20\% faster than \BagOfWords{} (with SIMD instructions).

\begin{table}
\caption{Inference speed per examples of the different MART models on the proposed v-QS and VPred implementation.}
\label{table:inference_speed}
\begin{tabular}{@{}lrr@{}}
\toprule
Method & \multicolumn{1}{l}{Extended v-QS (µs)} & \multicolumn{1}{l}{VPred  (µs)} \\ \midrule
\MARTGreedyMask & .636 & 8.65 \\
MART \BagOfWords & .754 & 15.4 \\
\MARTCatCART \Shingling & 2.80 & 2.16 \\
MART {\small \Shingling+\TargetMean} & 1.59 & 4.65 \\
MART \Shingling+\OneHot & 2.85 & 2.18 \\ \bottomrule
\end{tabular}
\end{table}

\section{Conclusion} \label{sec:conclusion}

In this work, we proposed a novel algorithm that enables decision forests to consume categorical-set features, effectively allowing them to model text without a need for feature transformation. Our solution equipped decision forests with the ability to efficiently find (greedy) splits in the space of sets of objects. We also extended QuickScorer, an inference algorithm to evaluate decision forests efficiently on modern CPUs, to include our proposed split.

Experiments on text classification showed that our method is competitive in terms of quality and inference speed compared to existing methods. Furthermore, an examination of the resulting models in terms of structure and sensitivity to our method's hyperparameter shows that our proposed method yields balanced trees and its performance is stable across various datasets and settings.

This work gives rise to a number of future research directions. Having established the feasibility of consuming raw textual features with decision forests, we are interested in variants of the proposed algorithm (e.g., ngram-based splits) and in better understanding their effect on different decision forest algorithms (RF vs. MART). Preventing the splitter from overfitting to training data, particularly on small datasets, is a topic worth exploring. Another question left unanswered is the interpretability of our proposed split: How one systematically assesses the role a particular term or split plays in the model needs to be studied.

\section{Acknowledgements}
We extend our thanks to Vytenis Sakenas, Dmitry Osmakov, Alexander Grushetsky, and the members of the RankLab team at Google for helpful discussions and insight. We also thank Masrour Zoghi for reviewing an early draft of this work.

\newpage
\balance
\bibliographystyle{ACM-Reference-Format}
\bibliography{main} 


\begin{thebibliography}{37}


\ifx \showCODEN    \undefined \def \showCODEN     #1{\unskip}     \fi
\ifx \showDOI      \undefined \def \showDOI       #1{#1}\fi
\ifx \showISBNx    \undefined \def \showISBNx     #1{\unskip}     \fi
\ifx \showISBNxiii \undefined \def \showISBNxiii  #1{\unskip}     \fi
\ifx \showISSN     \undefined \def \showISSN      #1{\unskip}     \fi
\ifx \showLCCN     \undefined \def \showLCCN      #1{\unskip}     \fi
\ifx \shownote     \undefined \def \shownote      #1{#1}          \fi
\ifx \showarticletitle \undefined \def \showarticletitle #1{#1}   \fi
\ifx \showURL      \undefined \def \showURL       {\relax}        \fi
\providecommand\bibfield[2]{#2}
\providecommand\bibinfo[2]{#2}
\providecommand\natexlab[1]{#1}
\providecommand\showeprint[2][]{arXiv:#2}

\bibitem[\protect\citeauthoryear{{Asadi}, {Lin}, and {de Vries}}{{Asadi}
  et~al\mbox{.}}{2014}]%
        {asadi2014}
\bibfield{author}{\bibinfo{person}{N. {Asadi}}, \bibinfo{person}{J. {Lin}},
  {and} \bibinfo{person}{A.~P. {de Vries}}.} \bibinfo{year}{2014}\natexlab{}.
\newblock \showarticletitle{Runtime Optimizations for Tree-Based Machine
  Learning Models}.
\newblock \bibinfo{journal}{\emph{IEEE Transactions on Knowledge and Data
  Engineering}} \bibinfo{volume}{26}, \bibinfo{number}{9}
  (\bibinfo{year}{2014}), \bibinfo{pages}{2281--2292}.
\newblock


\bibitem[\protect\citeauthoryear{Balestriero}{Balestriero}{2017}]%
        {balestriero2017}
\bibfield{author}{\bibinfo{person}{Randall Balestriero}.}
  \bibinfo{year}{2017}\natexlab{}.
\newblock \showarticletitle{Neural Decision Trees}.
\newblock \bibinfo{journal}{\emph{ArXiv}}  \bibinfo{volume}{abs/1702.07360}
  (\bibinfo{year}{2017}).
\newblock


\bibitem[\protect\citeauthoryear{Bengio, Ducharme, Vincent, and Janvin}{Bengio
  et~al\mbox{.}}{2003}]%
        {bengio2003}
\bibfield{author}{\bibinfo{person}{Yoshua Bengio}, \bibinfo{person}{R\'{e}jean
  Ducharme}, \bibinfo{person}{Pascal Vincent}, {and} \bibinfo{person}{Christian
  Janvin}.} \bibinfo{year}{2003}\natexlab{}.
\newblock \showarticletitle{A Neural Probabilistic Language Model}.
\newblock \bibinfo{journal}{\emph{Machine Learning Research}}
  \bibinfo{volume}{3}, \bibinfo{number}{null} (\bibinfo{date}{March}
  \bibinfo{year}{2003}), \bibinfo{pages}{1137–1155}.
\newblock
\showISSN{1532-4435}


\bibitem[\protect\citeauthoryear{Bloom}{Bloom}{1970}]%
        {bloom1970CACM}
\bibfield{author}{\bibinfo{person}{Burton~H. Bloom}.}
  \bibinfo{year}{1970}\natexlab{}.
\newblock \showarticletitle{Space/Time Trade-Offs in Hash Coding with Allowable
  Errors}.
\newblock \bibinfo{journal}{\emph{Commun. ACM}} \bibinfo{volume}{13},
  \bibinfo{number}{7} (\bibinfo{date}{July} \bibinfo{year}{1970}),
  \bibinfo{pages}{422–426}.
\newblock


\bibitem[\protect\citeauthoryear{Breiman}{Breiman}{1996}]%
        {breiman1996}
\bibfield{author}{\bibinfo{person}{Leo Breiman}.}
  \bibinfo{year}{1996}\natexlab{}.
\newblock \showarticletitle{Bagging predictors}.
\newblock \bibinfo{journal}{\emph{Machine Learning}}  \bibinfo{volume}{24}
  (\bibinfo{date}{Aug.} \bibinfo{year}{1996}), \bibinfo{pages}{123--140}.
\newblock


\bibitem[\protect\citeauthoryear{Breiman}{Breiman}{2001}]%
        {breiman2001}
\bibfield{author}{\bibinfo{person}{Leo Breiman}.}
  \bibinfo{year}{2001}\natexlab{}.
\newblock \showarticletitle{Random Forests}.
\newblock \bibinfo{journal}{\emph{Machine Learning}} \bibinfo{volume}{45},
  \bibinfo{number}{1} (\bibinfo{date}{Oct.} \bibinfo{year}{2001}),
  \bibinfo{pages}{5–32}.
\newblock


\bibitem[\protect\citeauthoryear{Breiman, Friedman, Olshen, and Stone}{Breiman
  et~al\mbox{.}}{1983}]%
        {breiman1983}
\bibfield{author}{\bibinfo{person}{Leo Breiman}, \bibinfo{person}{Joseph~H
  Friedman}, \bibinfo{person}{R.~A. Olshen}, {and} \bibinfo{person}{C.~J.
  Stone}.} \bibinfo{year}{1983}\natexlab{}.
\newblock \showarticletitle{Classification and Regression Trees}.
\newblock


\bibitem[\protect\citeauthoryear{Broder, Charikar, Frieze, and
  Mitzenmacher}{Broder et~al\mbox{.}}{2000}]%
        {broder2000Minhash}
\bibfield{author}{\bibinfo{person}{Andrei~Z Broder}, \bibinfo{person}{Moses
  Charikar}, \bibinfo{person}{Alan~M Frieze}, {and} \bibinfo{person}{Michael
  Mitzenmacher}.} \bibinfo{year}{2000}\natexlab{}.
\newblock \showarticletitle{Min-Wise Independent Permutations}.
\newblock \bibinfo{journal}{\emph{J. Comput. System Sci.}}
  \bibinfo{volume}{60}, \bibinfo{number}{3} (\bibinfo{year}{2000}),
  \bibinfo{pages}{630 -- 659}.
\newblock
\showISSN{0022-0000}


\bibitem[\protect\citeauthoryear{Bruch, Pfeifer, and Guillame-bert}{Bruch
  et~al\mbox{.}}{2020}]%
        {bruch2020learning}
\bibfield{author}{\bibinfo{person}{Sebastian Bruch}, \bibinfo{person}{Jan
  Pfeifer}, {and} \bibinfo{person}{Mathieu Guillame-bert}.}
  \bibinfo{year}{2020}\natexlab{}.
\newblock \bibinfo{title}{Learning Representations for Axis-Aligned Decision
  Forests through Input Perturbation}.
\newblock   (\bibinfo{year}{2020}).
\newblock
\showeprint[arxiv]{cs.LG/2007.14761}


\bibitem[\protect\citeauthoryear{Chen and Guestrin}{Chen and Guestrin}{2016}]%
        {xgboost2016KDD}
\bibfield{author}{\bibinfo{person}{Tianqi Chen} {and} \bibinfo{person}{Carlos
  Guestrin}.} \bibinfo{year}{2016}\natexlab{}.
\newblock \showarticletitle{XGBoost: A Scalable Tree Boosting System}. In
  \bibinfo{booktitle}{\emph{Proceedings of the 22nd ACM SIGKDD International
  Conference on Knowledge Discovery and Data Mining}}.
  \bibinfo{pages}{785–794}.
\newblock


\bibitem[\protect\citeauthoryear{Conneau and Kiela}{Conneau and Kiela}{2018}]%
        {senteval}
\bibfield{author}{\bibinfo{person}{Alexis Conneau} {and} \bibinfo{person}{Douwe
  Kiela}.} \bibinfo{year}{2018}\natexlab{}.
\newblock \showarticletitle{SentEval: An Evaluation Toolkit for Universal
  Sentence Representations}.
\newblock \bibinfo{journal}{\emph{arXiv preprint arXiv:1803.05449}}
  (\bibinfo{year}{2018}).
\newblock


\bibitem[\protect\citeauthoryear{Deng, Runger, Tuv, and Vladimir}{Deng
  et~al\mbox{.}}{2013}]%
        {deng2013}
\bibfield{author}{\bibinfo{person}{Houtao Deng}, \bibinfo{person}{George
  Runger}, \bibinfo{person}{Eugene Tuv}, {and} \bibinfo{person}{Martyanov
  Vladimir}.} \bibinfo{year}{2013}\natexlab{}.
\newblock \showarticletitle{A time series forest for classification and feature
  extraction}.
\newblock \bibinfo{journal}{\emph{Information Sciences}}  \bibinfo{volume}{239}
  (\bibinfo{year}{2013}), \bibinfo{pages}{142 -- 153}.
\newblock
\showISSN{0020-0255}


\bibitem[\protect\citeauthoryear{Devlin, Chang, Lee, and Toutanova}{Devlin
  et~al\mbox{.}}{2019}]%
        {devlin2019}
\bibfield{author}{\bibinfo{person}{Jacob Devlin}, \bibinfo{person}{Ming-Wei
  Chang}, \bibinfo{person}{Kenton Lee}, {and} \bibinfo{person}{Kristina
  Toutanova}.} \bibinfo{year}{2019}\natexlab{}.
\newblock \showarticletitle{BERT: Pre-training of Deep Bidirectional
  Transformers for Language Understanding}. In
  \bibinfo{booktitle}{\emph{NAACL-HLT}}.
\newblock


\bibitem[\protect\citeauthoryear{Duchi, Hazan, and Singer}{Duchi
  et~al\mbox{.}}{2011}]%
        {Duchi2011AdaGrad}
\bibfield{author}{\bibinfo{person}{John Duchi}, \bibinfo{person}{Elad Hazan},
  {and} \bibinfo{person}{Yoram Singer}.} \bibinfo{year}{2011}\natexlab{}.
\newblock \showarticletitle{Adaptive Subgradient Methods for Online Learning
  and Stochastic Optimization}.
\newblock \bibinfo{journal}{\emph{Machine Learning Research}}
  \bibinfo{volume}{12} (\bibinfo{date}{July} \bibinfo{year}{2011}),
  \bibinfo{pages}{2121--2159}.
\newblock
\showISSN{1532-4435}


\bibitem[\protect\citeauthoryear{Feng, Yu, and Zhou}{Feng
  et~al\mbox{.}}{2018}]%
        {feng2018}
\bibfield{author}{\bibinfo{person}{Ji Feng}, \bibinfo{person}{Yang Yu}, {and}
  \bibinfo{person}{Zhi-Hua Zhou}.} \bibinfo{year}{2018}\natexlab{}.
\newblock \showarticletitle{Multi-Layered Gradient Boosting Decision Trees}. In
  \bibinfo{booktitle}{\emph{NeurIPS}}.
\newblock


\bibitem[\protect\citeauthoryear{Friedman}{Friedman}{2001}]%
        {friedman2001}
\bibfield{author}{\bibinfo{person}{Jerome~H. Friedman}.}
  \bibinfo{year}{2001}\natexlab{}.
\newblock \showarticletitle{Greedy function approximation: A gradient boosting
  machine}.
\newblock \bibinfo{journal}{\emph{Ann. Statist.}} \bibinfo{volume}{29},
  \bibinfo{number}{5} (\bibinfo{date}{10} \bibinfo{year}{2001}),
  \bibinfo{pages}{1189--1232}.
\newblock


\bibitem[\protect\citeauthoryear{Ganjisaffar, Caruana, and Lopes}{Ganjisaffar
  et~al\mbox{.}}{2011}]%
        {ganjisaffar:2011:bagging}
\bibfield{author}{\bibinfo{person}{Yasser Ganjisaffar}, \bibinfo{person}{Rich
  Caruana}, {and} \bibinfo{person}{Cristina~Videira Lopes}.}
  \bibinfo{year}{2011}\natexlab{}.
\newblock \showarticletitle{Bagging Gradient-Boosted Trees for High Precision,
  Low Variance Ranking Models}. In \bibinfo{booktitle}{\emph{Proceedings of the
  34th International ACM SIGIR Conference on Research and Development in
  Information Retrieval}}. \bibinfo{pages}{85–94}.
\newblock


\bibitem[\protect\citeauthoryear{Geurts, Ernst, and Wehenkel}{Geurts
  et~al\mbox{.}}{2006}]%
        {geurts2006}
\bibfield{author}{\bibinfo{person}{Pierre Geurts}, \bibinfo{person}{Damien
  Ernst}, {and} \bibinfo{person}{Louis Wehenkel}.}
  \bibinfo{year}{2006}\natexlab{}.
\newblock \showarticletitle{Extremely Randomized Trees}.
\newblock \bibinfo{journal}{\emph{Machine Learning}} \bibinfo{volume}{63},
  \bibinfo{number}{1} (\bibinfo{date}{April} \bibinfo{year}{2006}),
  \bibinfo{pages}{3–42}.
\newblock
\showISSN{0885-6125}


\bibitem[\protect\citeauthoryear{Guillame-Bert and Dubrawski}{Guillame-Bert and
  Dubrawski}{2017}]%
        {guillamebert2017}
\bibfield{author}{\bibinfo{person}{Mathieu Guillame-Bert} {and}
  \bibinfo{person}{Artur Dubrawski}.} \bibinfo{year}{2017}\natexlab{}.
\newblock \showarticletitle{Classification of Time Sequences using Graphs of
  Temporal Constraints}.
\newblock \bibinfo{journal}{\emph{Journal of Machine Learning Research}}
  \bibinfo{volume}{18}, \bibinfo{number}{121} (\bibinfo{year}{2017}),
  \bibinfo{pages}{1--34}.
\newblock


\bibitem[\protect\citeauthoryear{He, Zhang, Ren, and Sun}{He
  et~al\mbox{.}}{2016}]%
        {he2016}
\bibfield{author}{\bibinfo{person}{Kaiming He}, \bibinfo{person}{Xiangyu
  Zhang}, \bibinfo{person}{Shaoqing Ren}, {and} \bibinfo{person}{Jian Sun}.}
  \bibinfo{year}{2016}\natexlab{}.
\newblock \showarticletitle{Deep Residual Learning for Image Recognition}.
  \bibinfo{pages}{770--778}.
\newblock


\bibitem[\protect\citeauthoryear{Jin and Agrawal}{Jin and Agrawal}{2003}]%
        {jin03}
\bibfield{author}{\bibinfo{person}{Ruoming Jin} {and} \bibinfo{person}{Gagan
  Agrawal}.} \bibinfo{year}{2003}\natexlab{}.
\newblock \showarticletitle{Communication and Memory Efficient Parallel
  Decision Tree Construction}. In \bibinfo{booktitle}{\emph{In Proceedings of
  Third SIAM Conference on Data Mining}}.
\newblock


\bibitem[\protect\citeauthoryear{Ke, Meng, Finley, Wang, Chen, Ma, Ye, and
  Liu}{Ke et~al\mbox{.}}{2017}]%
        {lightgbm2017nips}
\bibfield{author}{\bibinfo{person}{Guolin Ke}, \bibinfo{person}{Qi Meng},
  \bibinfo{person}{Thomas Finley}, \bibinfo{person}{Taifeng Wang},
  \bibinfo{person}{Wei Chen}, \bibinfo{person}{Weidong Ma},
  \bibinfo{person}{Qiwei Ye}, {and} \bibinfo{person}{Tie-Yan Liu}.}
  \bibinfo{year}{2017}\natexlab{}.
\newblock \showarticletitle{LightGBM: A Highly Efficient Gradient Boosting
  Decision Tree}.
\newblock In \bibinfo{booktitle}{\emph{Advances in Neural Information
  Processing Systems 30}}. \bibinfo{pages}{3146--3154}.
\newblock


\bibitem[\protect\citeauthoryear{Ke, Xu, Zhang, Bian, and Liu}{Ke
  et~al\mbox{.}}{2019}]%
        {Ke2019DeepGBM}
\bibfield{author}{\bibinfo{person}{Guolin Ke}, \bibinfo{person}{Zhenhui Xu},
  \bibinfo{person}{Jia Zhang}, \bibinfo{person}{Jiang Bian}, {and}
  \bibinfo{person}{Tie-Yan Liu}.} \bibinfo{year}{2019}\natexlab{}.
\newblock \showarticletitle{DeepGBM: A Deep Learning Framework Distilled by
  GBDT for Online Prediction Tasks}. In \bibinfo{booktitle}{\emph{Proceedings
  of the 25th ACM SIGKDD International Conference on Knowledge Discovery and
  Data Mining}}. \bibinfo{pages}{384–394}.
\newblock


\bibitem[\protect\citeauthoryear{Kontschieder, Fiterau, Criminisi, and
  Bul\`{o}}{Kontschieder et~al\mbox{.}}{2016}]%
        {kontschieder2016}
\bibfield{author}{\bibinfo{person}{Peter Kontschieder},
  \bibinfo{person}{Madalina Fiterau}, \bibinfo{person}{Antonio Criminisi},
  {and} \bibinfo{person}{Samuel~Rota Bul\`{o}}.}
  \bibinfo{year}{2016}\natexlab{}.
\newblock \showarticletitle{Deep Neural Decision Forests}. In
  \bibinfo{booktitle}{\emph{Proceedings of the Twenty-Fifth International Joint
  Conference on Artificial Intelligence}}
  \emph{(\bibinfo{series}{IJCAI’16})}. \bibinfo{publisher}{AAAI Press},
  \bibinfo{pages}{4190–4194}.
\newblock
\showISBNx{9781577357704}


\bibitem[\protect\citeauthoryear{LeCun, Bengio, and Hinton}{LeCun
  et~al\mbox{.}}{2015}]%
        {lecun2015}
\bibfield{author}{\bibinfo{person}{Yann LeCun}, \bibinfo{person}{Y. Bengio},
  {and} \bibinfo{person}{Geoffrey Hinton}.} \bibinfo{year}{2015}\natexlab{}.
\newblock \showarticletitle{Deep Learning}.
\newblock \bibinfo{journal}{\emph{Nature}}  \bibinfo{volume}{521}
  (\bibinfo{date}{05} \bibinfo{year}{2015}), \bibinfo{pages}{436--44}.
\newblock


\bibitem[\protect\citeauthoryear{Li, Qin, Wang, and Metzler}{Li
  et~al\mbox{.}}{2019}]%
        {Li2019CombiningDTandNNs}
\bibfield{author}{\bibinfo{person}{Pan Li}, \bibinfo{person}{Zhen Qin},
  \bibinfo{person}{Xuanhui Wang}, {and} \bibinfo{person}{Donald Metzler}.}
  \bibinfo{year}{2019}\natexlab{}.
\newblock \showarticletitle{Combining Decision Trees and Neural Networks for
  Learning-to-Rank in Personal Search}. In
  \bibinfo{booktitle}{\emph{Proceedings of the 25th ACM SIGKDD International
  Conference on Knowledge Discovery and Data Mining}}.
  \bibinfo{pages}{2032–2040}.
\newblock


\bibitem[\protect\citeauthoryear{Lucchese, Nardini, Orlando, Perego,
  Tonellotto, and Venturini}{Lucchese et~al\mbox{.}}{2016}]%
        {lucchese2016}
\bibfield{author}{\bibinfo{person}{Claudio Lucchese},
  \bibinfo{person}{Franco~Maria Nardini}, \bibinfo{person}{Salvatore Orlando},
  \bibinfo{person}{Raffaele Perego}, \bibinfo{person}{Nicola Tonellotto}, {and}
  \bibinfo{person}{Rossano Venturini}.} \bibinfo{year}{2016}\natexlab{}.
\newblock \showarticletitle{Exploiting CPU SIMD Extensions to Speed-up Document
  Scoring with Tree Ensembles}. \bibinfo{pages}{833--836}.
\newblock


\bibitem[\protect\citeauthoryear{Lucchese, Nardini, Orlando, Perego,
  Tonellotto, and Venturini}{Lucchese et~al\mbox{.}}{2017}]%
        {lucchese2017}
\bibfield{author}{\bibinfo{person}{Claudio Lucchese},
  \bibinfo{person}{Franco~Maria Nardini}, \bibinfo{person}{Salvatore Orlando},
  \bibinfo{person}{Raffaele Perego}, \bibinfo{person}{Nicola Tonellotto}, {and}
  \bibinfo{person}{Rossano Venturini}.} \bibinfo{year}{2017}\natexlab{}.
\newblock \bibinfo{booktitle}{\emph{QuickScorer: Efficient Traversal of Large
  Ensembles of Decision Trees}}.
\newblock \bibinfo{pages}{383--387}.
\newblock
\showISBNx{978-3-319-71272-7}


\bibitem[\protect\citeauthoryear{Manning, Raghavan, and Sch\"{u}tze}{Manning
  et~al\mbox{.}}{2008}]%
        {manning2008}
\bibfield{author}{\bibinfo{person}{Christopher~D. Manning},
  \bibinfo{person}{Prabhakar Raghavan}, {and} \bibinfo{person}{Hinrich
  Sch\"{u}tze}.} \bibinfo{year}{2008}\natexlab{}.
\newblock \bibinfo{booktitle}{\emph{Introduction to Information Retrieval}}.
\newblock \bibinfo{publisher}{Cambridge University Press}.
\newblock
\showISBNx{0521865719}


\bibitem[\protect\citeauthoryear{Mikolov, Sutskever, Chen, Corrado, and
  Dean}{Mikolov et~al\mbox{.}}{2013}]%
        {mikolov2013}
\bibfield{author}{\bibinfo{person}{Tomas Mikolov}, \bibinfo{person}{Ilya
  Sutskever}, \bibinfo{person}{Kai Chen}, \bibinfo{person}{G.s Corrado}, {and}
  \bibinfo{person}{Jeffrey Dean}.} \bibinfo{year}{2013}\natexlab{}.
\newblock \showarticletitle{Distributed Representations of Words and Phrases
  and their Compositionality}.
\newblock \bibinfo{journal}{\emph{Advances in Neural Information Processing
  Systems}}  \bibinfo{volume}{26} (\bibinfo{date}{10} \bibinfo{year}{2013}).
\newblock


\bibitem[\protect\citeauthoryear{Pennington, Socher, and Manning}{Pennington
  et~al\mbox{.}}{2014}]%
        {pennington2014}
\bibfield{author}{\bibinfo{person}{Jeffrey Pennington},
  \bibinfo{person}{Richard Socher}, {and} \bibinfo{person}{Christopher
  Manning}.} \bibinfo{year}{2014}\natexlab{}.
\newblock \showarticletitle{{G}lo{V}e: Global Vectors for Word Representation}.
  In \bibinfo{booktitle}{\emph{Proceedings of the 2014 Conference on Empirical
  Methods in Natural Language Processing ({EMNLP})}}.
  \bibinfo{publisher}{Association for Computational Linguistics},
  \bibinfo{address}{Doha, Qatar}, \bibinfo{pages}{1532--1543}.
\newblock


\bibitem[\protect\citeauthoryear{Prokhorenkova, Gusev, Vorobev, Dorogush, and
  Gulin}{Prokhorenkova et~al\mbox{.}}{2018}]%
        {catboost}
\bibfield{author}{\bibinfo{person}{Liudmila Prokhorenkova},
  \bibinfo{person}{Gleb Gusev}, \bibinfo{person}{Aleksandr Vorobev},
  \bibinfo{person}{Anna~Veronika Dorogush}, {and} \bibinfo{person}{Andrey
  Gulin}.} \bibinfo{year}{2018}\natexlab{}.
\newblock \showarticletitle{CatBoost: unbiased boosting with categorical
  features}.
\newblock In \bibinfo{booktitle}{\emph{Advances in Neural Information
  Processing Systems 31}}, \bibfield{editor}{\bibinfo{person}{S.~Bengio},
  \bibinfo{person}{H.~Wallach}, \bibinfo{person}{H.~Larochelle},
  \bibinfo{person}{K.~Grauman}, \bibinfo{person}{N.~Cesa-Bianchi}, {and}
  \bibinfo{person}{R.~Garnett}} (Eds.). \bibinfo{publisher}{Curran Associates,
  Inc.}, \bibinfo{pages}{6638--6648}.
\newblock


\bibitem[\protect\citeauthoryear{Rashmi and Gilad-Bachrach}{Rashmi and
  Gilad-Bachrach}{2015}]%
        {rashmi2015}
\bibfield{author}{\bibinfo{person}{K. Rashmi} {and} \bibinfo{person}{Ran
  Gilad-Bachrach}.} \bibinfo{year}{2015}\natexlab{}.
\newblock \showarticletitle{DART: Dropouts meet Multiple Additive Regression
  Trees}.
\newblock  (\bibinfo{date}{05} \bibinfo{year}{2015}).
\newblock


\bibitem[\protect\citeauthoryear{Rodríguez and Alonso}{Rodríguez and
  Alonso}{2004}]%
        {rodriguez2004}
\bibfield{author}{\bibinfo{person}{Juan Rodríguez} {and}
  \bibinfo{person}{Carlos Alonso}.} \bibinfo{year}{2004}\natexlab{}.
\newblock \showarticletitle{Interval and dynamic time warping-based decision
  trees}. \bibinfo{pages}{548--552}.
\newblock


\bibitem[\protect\citeauthoryear{Wang, Girshick, Gupta, and He}{Wang
  et~al\mbox{.}}{2018}]%
        {wang2018}
\bibfield{author}{\bibinfo{person}{Xiaolong Wang}, \bibinfo{person}{Ross~B.
  Girshick}, \bibinfo{person}{Abhinav Gupta}, {and} \bibinfo{person}{Kaiming
  He}.} \bibinfo{year}{2018}\natexlab{}.
\newblock \showarticletitle{Non-local Neural Networks}.
\newblock \bibinfo{journal}{\emph{2018 IEEE/CVF Conference on Computer Vision
  and Pattern Recognition}} (\bibinfo{year}{2018}),
  \bibinfo{pages}{7794--7803}.
\newblock


\bibitem[\protect\citeauthoryear{Ye, Zhou, Zou, Gao, and Zhang}{Ye
  et~al\mbox{.}}{2018}]%
        {ye2018}
\bibfield{author}{\bibinfo{person}{Ting Ye}, \bibinfo{person}{Hucheng Zhou},
  \bibinfo{person}{Will Zou}, \bibinfo{person}{Bin Gao}, {and}
  \bibinfo{person}{Ruofei Zhang}.} \bibinfo{year}{2018}\natexlab{}.
\newblock \showarticletitle{RapidScorer: Fast Tree Ensemble Evaluation by
  Maximizing Compactness in Data Level Parallelization}.
  \bibinfo{pages}{941--950}.
\newblock


\bibitem[\protect\citeauthoryear{Zaheer, Kottur, Ravanbakhsh, P{\'o}czos,
  Salakhutdinov, and Smola}{Zaheer et~al\mbox{.}}{2017}]%
        {zaheer2017}
\bibfield{author}{\bibinfo{person}{Manzil Zaheer}, \bibinfo{person}{Satwik
  Kottur}, \bibinfo{person}{Siamak Ravanbakhsh}, \bibinfo{person}{Barnab{\'a}s
  P{\'o}czos}, \bibinfo{person}{Ruslan Salakhutdinov}, {and}
  \bibinfo{person}{Alexander~J. Smola}.} \bibinfo{year}{2017}\natexlab{}.
\newblock \showarticletitle{Deep Sets}.
\newblock \bibinfo{journal}{\emph{ArXiv}}  \bibinfo{volume}{abs/1703.06114}
  (\bibinfo{year}{2017}).
\newblock


\end{thebibliography}

\end{document}